\documentclass[10pt, a4paper]{article}
\usepackage{float}

\usepackage{lrec-coling2024} 
\usepackage{tabularray}
\usepackage{soul}
\usepackage{placeins}
\usepackage{tcolorbox}
\usepackage{caption}
\usepackage{comment}
\usepackage{multirow}
\usepackage{graphicx}
\usepackage{authblk}

\usepackage{todonotes} 
\newcommand*{\affmark}[1][*]{\textsuperscript{#1}}

\usepackage{booktabs} 

\title{Can Large Language Models Automatically Score Proficiency of Written Essays?}

\name{Watheq Mansour\affmark[1]\textsuperscript{*}\thanks{*The work was done while at Qatar Univeristy.}, Salam Albatarni\affmark[2], Sohaila Eltanbouly\affmark[2], Tamer Elsayed\affmark[2]} 

\address{
\affmark[1]The University of Queensland, Brisbane, Australia \\
\affmark[2]Computer Science and Engineering Department, Qatar University, Doha, Qatar\\
w.mansour@uq.edu.au, 
         \{salam.albatarani, sohaila.eltanbouly, telsayed\}@qu.edu.qa\\
         }

\abstract{
Although several methods were proposed to address the problem of automated essay scoring (AES) in the last 50 years, there is still much to desire in terms of effectiveness. Large Language Models (LLMs) are transformer-based models that demonstrate extraordinary capabilities on various tasks. In this paper, we test the ability of LLMs, given their powerful linguistic knowledge, to analyze and effectively score written essays. We experimented with two popular LLMs, namely ChatGPT and Llama. We aim to check if these models can do this task and, if so, how their performance is positioned among the state-of-the-art (SOTA) models across two levels, holistically and per individual writing trait. 
We utilized prompt-engineering tactics in designing four different prompts to bring their maximum potential to this task. 
Our experiments conducted on the ASAP dataset revealed several interesting observations. First, choosing the right prompt depends highly on the model and nature of the task. Second, the two LLMs exhibited comparable average performance in AES, with a slight advantage for ChatGPT. Finally, despite the performance gap between the two LLMs and SOTA models in terms of predictions, they provide feedback to enhance the quality of the essays, which can potentially help both teachers and students.
 \\ \newline  }

\begin{document}

\maketitleabstract

\section{Introduction}
Automated Essay Scoring (AES), one of the earliest research problems in natural language processing~\citep{page1966imminence}, aims to develop automated tools for assessing the proficiency of written essays. 
It can save a huge amount of time spent by teachers assessing essays manually while providing free feedback to students. In addition to being more consistent over time, AES systems can deliver more impartial assessments than humans.

Over the past 50 years, a wide variety of approaches were proposed to address AES, including learning from hand-crafted features~\citep{phandi2015flexible, Chen2013AutomatedES}, and neural approaches including pre-trained language models~\citep{dong2017attention, wu2022beyond}. 
Most of them focus on predicting an overall holistic score that closely matches human judgment. Other studies focus on providing feedback by estimating quality scores on multiple traits of the essay, e.g., coherence~\citep{higgins2004evaluating}, relevance~\citep{louis2010off}, and organization~\citep{persing2010modeling} among others.

Recently, Large Language Models (LLMs) were introduced as cutting-edge transformer-based models demonstrating impressive capabilities on various tasks \citep{meoni-etal-2023-large, kocmi2023large}. In this paper, we check if the language proficiency retained by LLMs (represented by ChatGPT\footnote{\url{https://chat.openai.com/chat}} and Llama 2\footnote{\url{https://ai.meta.com/llama/}} in our study) is useful in scoring essays. 
We test the capabilities of LLMs in this task across two levels, holistically and per individual writing trait. 
We also compare its performance to current state-of-the-art (SOTA) models. Furthermore, we analyze its performance and check its consistency in scoring the same essays.  

To this end, our aim in this work is to answer the following research questions: 
\begin{itemize}
\item \textbf{RQ1:} What is the effect of prompt engineering on the performance of LLMs for AES?
\item \textbf{RQ2:} Is the performance of LLMs for AES consistent across prompts? 
\item \textbf{RQ3:} How far is the performance of ChatGPT and Llama from SOTA, holistically and per trait?
\end{itemize}

We carried out the experiments on the full Automated Student Assessment Prize (ASAP) dataset\footnote{\label{ASAP_note}\url{https://www.kaggle.com/c/asap-aes}} comprising 8 tasks and 12978 essays. 
We designed four different prompts in an incremental fashion, where each adds more 
information to the prior one, and collected its responses for all essays. We also recorded the predicted scores and feedback for each essay and released them to the research community for further analysis.

Our developed prompts led to notable enhancements in the performance of LLMs, reaching a peak Quadratic Weighted Kappa~\cite{cohen1968weighted} score of 0.606 and 0.562 by ChatGPT and Llama 2, respectively. Although those scores lag behind SOTA models with respect to scoring prediction, LLMs have an advantage in terms of generated feedback. For all the experimented essays, they provided feedback to enhance their quality.

Our contribution in this work is four-fold: 

\begin{itemize}
\item We evaluate the predictive performance of two LLMs, namely ChatGPT and Llama 2, for AES at the holistic and trait-based levels and compare it against the SOTA models.
\item We show the impact of prompt engineering on ChatGPT's and Llama's performance by designing four different prompts with incremental task elaboration.  
\item We study the consistency of ChatGPT and Llama 2 in AES.
\item We make all responses of the LLMs (including their feedback) publicly available to enable future analysis and research.\footnote{\url{https://github.com/Watheq9/AES-with-LLMs}}
\end{itemize} 

The rest of the paper is organized as follows. Section \ref{related_work} discusses the work related to AES and LLMs. The detailed description of our experiment and prompt designs are provided in Section \ref{methodology}. Section \ref{results} presents the experimental results, the analysis of prompt engineering, and performance consistency. Finally, we conclude in Section \ref{conclusion}.

\section{Related Work} \label{related_work}

Previous attempts to address the AES problem relied either on crafting features to represent essays, syntactically~\cite{yannakoudakis2011new} or lexicallay~\cite{Chen2013AutomatedES, phandi2015flexible}, or adopting neural networks~\cite{xie2022automated}. As a matter of fact, SOTA performance was achieved using a set of Multi-layer perceptrons that simultaneously conduct regression and ranking optimization \cite{xie2022automated}.

Following the advancement of neural networks, LLMs were utilized in various applications. Consequently, a handful of researchers explored fine-tuning LLMs, e.g., BERT~\cite{devlin2018bert}, for the AES task, which yielded a remarkable performance~\cite{wang2022use, yang2020enhancing}.

ChatGPT and Llama\footnote{Hereafter we will refer to Llama 2 as Llama.} were built using reinforcement learning from human feedback to capture contextual meanings in a much deeper way, leading to a better language generation. Those models can be utilized for multiple tasks, e.g., code generation and text paraphrasing.
Although work utilizing LLMs for AES is limited, the field is quickly evolving. Existing work solely focus on proprietary models. For example, \citet{mizumoto2023exploring} utilized OpenAI’s text-davinci-003 model for text completion to automatically score the essays of the TOEFL11 dataset and employed the generated scores as features for a Bayesian regression model. \citet{naismith-etal-2023-automated} tested GPT-4~\cite{achiam2023gpt} on assessing the discourse coherence of essays, yielding promising results. Similarly, \citet{yancey-etal-2023-rating} focused on GPT-3.5 and GPT-4 with Common European Framework of Reference for Languages (CEFR) short essays dataset. Both models fell short behind a simple XGBoost baseline.

Although prior research highlighted the underperformance of LLMs, some gaps persist. Firstly, existing studies focus solely on OpenAI's paid models. Secondly, none of them have tested these models on the ASAP dataset, an essential benchmark in AES for comparing with SOTA models.

In this paper, we aim to assess the capability of using LLMs in AES over two levels, holistically and per trait. Moreover, different prompts are crafted to study the effect of prompt engineering on the performance of LLMs.

\section{Methodology of Study} \label{methodology}

This section presents the outline of the adopted models and dataset, the design of the experimented prompts, and the experimental design.

\subsection{Dataset and Models}

We used the ASAP dataset, which is broadly used to evaluate AES systems \citep{ke2019automated}. ASAP was proposed in a competition on Kaggle in 2012. It contains 8 different tasks written in English by students ranging from grades 7 to 10. 
There are three types of tasks: persuasive, source-dependent response, and narrative. Table~\ref{tab:ASAP} describes the properties of each task.

We have chosen two LLM models for our experiments, namely, 
ChatGPT and Llama. ChatGPT has demonstrated remarkable capabilities on various tasks and gained significant attention across diverse domains. In our experiments, we utilized the ChatGPT API, namely gpt-3.5-turbo-0301 (a Snapshot from March 1st, 2023).\footnote{\url{https://platform.openai.com/docs/models}} This model is optimized for dialogue, and its performance is on par with Instruct Davinci. 
Nevertheless, ChatGPT API is not open source, which imposes scalability limitations due to its associated cost.\footnote{The total cost of the experiments with ChatGPT in this work amounted to approximately \$120.} Alternatively, Llama is an open-source LLM developed by Meta. It outperforms other open-source language models (e.g., Falcon and MPT) on different benchmarks~\cite{llama-paper}. In this work, we used Llama-2-13b-chat-hf.\footnote{\url{https://huggingface.co/meta-llama/Llama-2-13b-chat-hf}} For both models, the temperature is set to 0 to guarantee reproducible results.
\\

\begin{table}[t]
\centering
\footnotesize
\setlength{\tabcolsep}{4pt}
\begin{tblr}{
    colspec = {cp{0.9cm}cp{0.8cm}c},
  cells = {c},
  cell{2}{2} = {r=2}{},
  cell{4}{2} = {r=4}{},
  cell{8}{2} = {r=2}{},
  hline{1-2,4,8,10} = {-}{},
}
\centering Task & Essay Type  & Avg. length & Score Range & \# Essays \\
1    & Persuasive  & 350            & 2-12        & 1783             \\
2    &             & 350            & 1-6         & 1800             \\
3    & Source-Dependent           & 150            & 0-3         & 1726             \\
4    &             & 150            & 0-3         & 1772             \\
5    &  & 150            & 0-4         & 1805             \\
6    &             & 150            & 0-4         & 1800             \\
7    & Narrative   & 250            & 0-30        & 1569             \\
8    &             & 650            & 0-60        & 723              
\end{tblr}
\caption{Properties of the different tasks in the ASAP dataset.}
\label{tab:ASAP}
\end{table}

\begin{minipage}[t]{\linewidth}
\begin{tcolorbox}[colback=gray!10, colframe=black, boxrule=0.5pt, sharp corners, left=2pt, right=2pt, top=2pt, bottom=2pt]
{\fontsize{9}{12}\selectfont Note that the essay might contain some anonymization tokens that are placeholders added to anonymize sensitive information and ensure privacy. They replace personal names, locations, numbers, times, dates, organizations, and other specific information that could identify individuals or institutions. The anonymization tokens are: @PERSON\#, @LOCATION\#, @NUM\#, @TIME, @MONTH, @DATE, and @CAPS\#, where \# refers to a number.
}
\end{tcolorbox}
\captionof{figure}{The preprocessing instructions.}
\label{fig:preprocessing_instruction}
\end{minipage}

\subsection{Prompt Design} 

Four prompts are designed to assess the performance of ChatGPT and Llama. The prompts are structured in an incremental fashion, with each subsequent prompt building upon the preceding one by adding extra task elaboration. The key inputs in all the prompts are the task (i.e., the prompt the essay was written in response to) and the essay. Then, the scoring rubric\footnote{Scoring guidelines that specify each score criteria} and one-shot example are added to the subsequent prompts. 

\begin{figure}[H]
\centering
\includegraphics[width=0.5\textwidth]{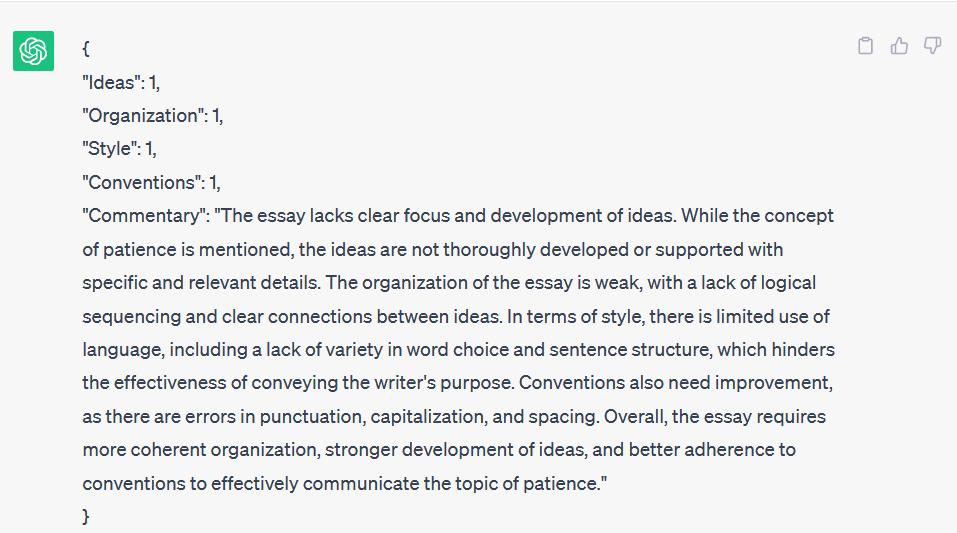}
\caption{A sample of ChatGPT response for essay with id 17834 from task 7 using prompt D.} \label{fig:chatgpt_response}
\end{figure}

Ensuring the clarity of the prompt given to the LLMs is vital.
Therefore, we follow some prompt-engineering tactics advised by OpenAI\footnote{\url{https://platform.openai.com/docs/guides/gpt-best-practices/strategy-write-clear-instructions}} to enhance the prompt comprehension. 
First, proper delimiters are used to distinguish the inputs (essay, rubrics, and one-shot examples) from the instruction text, preventing prompt injection issues.
Second, the prompt is designed as a series of sequential instructions, pushing the models to complete the steps in order. 
Third, we instruct the models to preprocess the input text (as shown in Figure~\ref{fig:preprocessing_instruction}), 
as certain placeholders are present within the ASAP essays. Finally, we ask for a JSON output structure specifying the desired outputs. Overall, we designed four prompt variants that meet these guidelines as follows:

\begin{itemize} 
  \item {\bfseries Prompt A} (shown in Figures~\ref{prompt1_structure} and \ref{prompt1_structure_llama}) is a simple prompt asking to evaluate the input essay \emph{without defining the rubrics}. It anchors our expectations on the performance of ChatGPT and Llama without any extra information.   
\begin{minipage}[b]{\linewidth}
\begin{tcolorbox}[colback=gray!10, colframe=black, boxrule=0.5pt, sharp corners, left=2pt, right=2pt, top=2pt, bottom=2pt]
{\fontsize{9}{12}\selectfont Your job is to evaluate the provided essay on a scale from 1 to 6.

\{preprocessing\_instruction\}

Here is a summary of the required steps:

1. Evaluate the essay (after suitable replacement of the anonymization tokens wherever needed) on a scale from 1 to 6.

2. Format your response in JSON as follows:

\{

``Total\_score'': your evaluation score, 

``Commentary'': your feedback on the essay

\}

Here is the prompt delimited by <>: <Prompt text> 

And here is the essay, which you need to evaluate, delimited by triple backticks: ```input\_essay```
}
\end{tcolorbox}
\captionof{figure}{Prompt A, a simple prompt asking ChatGPT to score an essay according to the score range.}
\label{prompt1_structure}
\end{minipage}

\begin{minipage}[b]{\linewidth}
\begin{tcolorbox}[colback=gray!10, colframe=black, boxrule=0.5pt, sharp corners, left=2pt, right=2pt, top=2pt, bottom=2pt]
{\fontsize{9}{12}\selectfont You will be given a prompt and an essay that was written in response to that prompt.
Your job is to score the provided essay on a scale from 1 to 6.

\{preprocessing\_instruction\}

Here is the prompt delimited by <>: <task\_prompt>

And here is the essay, which you need to score, delimited by triple backticks: ```input\_essay```

Format your response in JSON as follows:

\{

``Total\_score'': your evaluation score,

\}

The response should include only the JSON format.
}
\end{tcolorbox}
\captionof{figure}{Prompt A, a simple prompt asking Llama to score an essay according to score range.}
\label{prompt1_structure_llama}
\end{minipage}
  
  \item {\bfseries Prompt B} \emph{adds the rubric guidelines} to prompt A, allowing the models to understand the task more. We utilized the rubrics attached to each task in ASAP, which include definitions for each score point in the scale. The rubric assists in providing an objective framework to assess the quality and fulfillment of specific criteria.
  
  \item {\bfseries Prompt C} expands on Prompt B by \emph{adopting a one-shot example strategy}. This prompt will demonstrate whether ChatGPT and Llama are able to learn from the provided example and enhance its performance. 

  \item {\bfseries Prompt D} (shown in Figures~\ref{prompt4_structure}, \ref{prompt4_llama_history}, and \ref{prompt4_llama}) \emph{formats prompt C as a chat between a system, an assistant, and a user}, dividing the input and output into suitable roles.\footnote{Such roles are defined by the LLM.} Thus, the expected input and output hopefully become much clearer to the LLM. Figure~\ref{fig:chatgpt_response} presents an example of ChatGPT's response to an essay from Task 7 using this prompt.
\end{itemize}

\begin{figure}[H]
\begin{tcolorbox}[colback=gray!10, colframe=black, boxrule=0.5pt, sharp corners, left=0.5pt, right=0.5pt, top=0.5pt, bottom=0.5pt]
{\fontsize{9}{12}\selectfont 

\{``role'': ``system'',

``content'': “You are a helpful pattern-following assistant that evaluates essays written by students in response to a given prompt. Your job is to evaluate the provided essay per the following Rubric Guidelines delimited by <>: <\{rubric\_guidelines\}>

\{preprocessing\_instruction\}

Here is a summary of the required steps

1. Evaluate the essay (after suitable replacement of the anonymization tokens wherever needed) per the described rubric guidelines.

2. Format your response in JSON as follows:

\{

``Total\_score'': your evaluation score, 

``Commentary'': your feedback on the essay

\}

Here is the prompt delimited by <>: <{task\_prompt}> '''
\},

\{ ``role'': ``system'',

``name'': ``example\_user'',

``content'': example\_input\},

\{``role'': ``system'',

``name'': ``example\_assistant'',

``content'': example\_output\},

\{``role'': ``user'',

 ``content'': actual\_input,\}
}
\end{tcolorbox}
\captionof{figure}{Prompt D for ChatGPT, formats prompt C as a chat between a system an assistant, and a user.}
\label{prompt4_structure}
\end{figure}

The prompts for ChatGPT and Llama contain the same instructions but in different formats. We tried multiple arrangements of the instructions inside the prompt to see the best instruction order for each LLM. 
After several trials with the order of the instructions inside the prompts, we noticed two main differences. First, ChatGPT benefits from adding a concise list of required steps, whereas Llama did not exhibit similar enhancements from these steps. 
Second, ChatGPT provided the required JSON format, which was one of the required steps, regardless of the position of this requirement inside the prompt. On the other hand, with Llama, the required format is only generated when explicitly specify it at the very end of the prompt. 
Another difference in the prompts between ChatGPT and Llama is that in Llama's prompts, we did not ask for feedback. We did this to keep the responses concise because Llama often offers extra feedback even when it is not asked to provide one. Figures \ref{prompt1_structure} and \ref{prompt1_structure_llama} show the structure of prompt A for ChatGPT and Llama. Prompts B and C have similar formats with the addition of the rubric and the one-shot example.

For Prompt D, ChatGPT has clear and well-defined roles for formatting the prompt as a chat. Figure \ref{prompt4_structure} shows prompt D for ChatGPT. Similarly, Llama has the capability to provide system instruction but with a different format. Llama has an advantage that it can store the history of the messages. So we utilized this by adding the system instructions and the one-shot example as history, and the prompt contained only the input essay. However, we noticed Llama tends to forget the constraint of requiring the response to be in JSON format, so we repeated this instruction with each input. Figures \ref{prompt4_llama_history} and \ref{prompt4_llama} show the history and prompt D for Llama.

\begin{figure}[H]
\begin{tcolorbox}[colback=gray!10, colframe=black, boxrule=0.5pt, sharp corners, left=0.5pt, right=0.5pt, top=0.5pt, bottom=0.5pt]
{\fontsize{9}{12}\selectfont 

[
`system', 'You are a helpful pattern-following assistant that evaluates essays written by students in response to a given prompt.
Your job is to score the provided essay per the following Rubric Guidelines delimited by <>: <\{rubric\_guidelines\}>

\{preprocessing\_instruction\}

Here is the prompt delimited by <>:  <{task\_prompt}>

Format your response in JSON as follows:

\{

``Total\_score'': your evaluation score,

\}

The response should include only the JSON format.

], 

[
'[INST]', 'example\_input'
], 

[
'[/INST]', '\{example\_output\}'
]
}

\end{tcolorbox}
\captionof{figure}{The history message provided to Llama for Prompt D. In Llama [INST] indicates the user input, and [/INST] indicates Llama's response.}
\label{prompt4_llama_history}
\end{figure}

\begin{figure}[H]
\begin{tcolorbox}[colback=gray!10, colframe=black, boxrule=0.5pt, sharp corners, left=0.5pt, right=0.5pt, top=0.5pt, bottom=0.5pt]
{\fontsize{9}{12}\selectfont 

\{input\_essay\}

The response should include only the JSON format.

}

\end{tcolorbox}
\captionof{figure}{ Prompt D for Llama. It contains the input essay only. All other instructions are included in the conversation history.}
\label{prompt4_llama}
\end{figure}

\subsection{Experiment Design}
Considering ChatGPT's paid API, testing 4 prompts on all tasks is quite costly. Given that ASAP tasks are divided into three categories, we performed the experiments in two phases for both ChatGPT and Llama. In the \textbf{first phase}, we experiment with one task from each category on all of our prompts to choose the best prompt for each category. In the \textbf{second phase}, we apply that prompt to the remaining tasks in that category. From each category, we opt for the task with the shortest overall length (rubric guidelines, essay resource, and task question). So, we selected task 1 (persuasive), task 5 (source-dependent), and task 7 (narrative) for the first phase.

\section{Experimental Results and Discussion} \label{results}

In this Section, we discuss the performance of ChatGPT and Llama for AES. Moreover, we provide a thorough examination of the differences between ChatGPT and Llama when it comes to judging essay proficiency and their consistency. Recall that in this work, we aim to answer the following research questions: 
\begin{itemize}
\item \textbf{RQ1:} What is the effect of prompt engineering on the performance of LLMs for AES? 
\item \textbf{RQ2:} Is the performance of LLMs for AES consistent across prompts? 
\item \textbf{RQ3:} How far is the performance of ChatGPT and Llama from SOTA, holistically and per trait?
\end{itemize}

To evaluate the AES performance of ChatGPT and Llama, we use Quadratic Weighted Kappa (QWK) \cite{cohen1968weighted}, which is a widely used measure for AES. 
It is worth noting that, for ASAP tasks 7 and 8, we computed the holistic scores out of the predicted trait scores per the formula mentioned in the corresponding task guidelines. 

Our \textbf{baselines} are SOTA on both the holistic and trait levels. As such, we contrast our LLM models with~\citet{xie2022automated} and \citet{jiang-etal-2023-improving}, which achieved SOTA performance for in-prompt AES and cross-prompt AES respectively, from the holistic perspective. We also compare our LLM models against~\citet{mathias_bhattacharyya_2020_neural} over the traits level.

\subsection{Effect of Prompt Engineering (RQ1)}
\label{rq1}

\begin{table*}[h]
\center
\begin{tabular}{llcccc}
\hline
\multirow{2}{*}{LLM} & \multirow{2}{*}{Task} & \multicolumn{4}{c}{Prompt} \\
\cmidrule(lr){3-6} &  & A & B & C & D \\ \hline
ChatGPT & Task 1 (Persuasive) & 0.057 & 0.080 & \textbf{0.120} & 0.113 \\
 & Task 5 (Source-Dep.) & 0.552 & \textbf{0.576} & 0.476 & 0.455 \\
 & Task 7 (Narrative) & 0.060 & 0.027 & 0.055 & \textbf{0.123} \\ \hline
Llama & Task 1 (Persuasive) & \textbf{0.331} & 0.092 & 0.154 & 0.170 \\
 & Task 5 (Source-Dep.) & 0.047 & 0.346 & \textbf{0.418} & 0.180 \\
 & Task 7 (Narrative) & 0.268 & \textbf{0.359} & 0.037 & 0.224 \\ \hline
\end{tabular}%
\caption{Performance of LLMs measured in QWK over different prompt designs per task category. The best performance in each row (model and task category) is boldfaced.}
\label{tab:results}
\end{table*}

Table~\ref{tab:results} presents the results of the first phase of our experiments, which reveal several observations.

\paragraph{Effect on ChatGPT} For ChatGPT, first, prompts B, C, and D exhibit better performance than prompt A (except for the source-dependant task, where prompt A outperforms prompts C and D). This is expected as they all provide more context that is apparently needed for ChatGPT to comprehend the task better. 
Second, the best prompt is different across different task categories. More specifically, prompts C, D, and B were the best for the persuasive, narrative, and source-dependent categories, respectively. This is a very interesting outcome that highlights the importance of prompt engineering for AES in general and AES different task categories in particular. Moreover, a closer look at the task definition of the source-dependent category explains why prompt B was the best; in such category, the source text is provided, and students were mandated to include evidence from it, making it a somewhat ``self-contained'' task and prompt B was just enough for that with no need for showing an example (as in prompts C and D). That also explains why the source-dependent category, represented by task 5, seems to be much easier (over all prompts) than the other categories. In contrast, the persuasive and narrative categories benefited from the one-shot setup of prompts C and D; however, both still exhibit low absolute performance.

\paragraph{Effect on Llama} On the other hand, Llama exhibits a totally different performance than ChatGPT. As for the persuasive task category, prompt~A achieved the best performance. Recall that in the argumentative/persuasive task, essays typically present evidence and facts to support their stated argument. Since Llama's pre-training data incorporates a more extensive corpus of factual information than others
~\cite{llama-paper}, it is possible that it grasps the task better with the basic instructions in prompt A, while the added context in the other prompts confuses the model and makes predictions more challenging.
As for the remaining tasks, adding the rubrics in prompt B improved the context clarity and the scoring criteria over prompt A. Moreover, it seems that adding a one-shot example appeared to add noise to the prompt for the narrative task, although it was useful in the source-dependent task. 

\paragraph{Overall} When observing both the performance of ChatGPT and Llama, we can notice a huge discrepancy. For example, Llama's performance varies notably between different prompts for the same task, whereas this behavior is not observed with ChatGPT. 

\subsection{Consistency (RQ2)} 
To show this discrepancy in more detail, we assessed the consistency of the scores generated by ChatGPT and Llama across the different prompts. As the prompts build upon one another, it is anticipated that the performance of the LLM will have a reasonable agreement between the various prompts because \emph{they all share the same core instructions}. To perform this analysis, we calculate the QWK score between each pair of prompts for tasks 1, 5, and 7 to assess the level of agreement among each LLM response across the different prompts. Table \ref{tab:consistency_evaluation} shows the results of this experiment for the 6 pairs of prompts. For ChatGPT, the adjacent pairs always have a higher QWK; for example, the A-B pairs have higher scores than A-C and A-D for the three tasks. This behavior is expected since the adjacent prompts have higher context similarity. The average QWK for ChatGPT for tasks 1, 5, and 7 are 0.62, 0.58, and 0.64. On the other hand, Llama showed a significant disagreement between the responses of the different prompts, with no observable increasing or decreasing trend. The average QWK is 0.4, 0.2, and 0.17 for tasks 1, 5, and 7, respectively. 

From this analysis, Llama demonstrated inconsistent performance in AES, which suggests that it is highly sensitive to prompt selection. ChatGPT showed higher agreement across the different prompts and tasks, indicating that ChatGPT exhibits consistent performance in AES. It is worth noting that in the ASAP dataset, the average agreement score between the two human raters is 0.75~\footnote{For rater1\_domain1 and rater2\_domain1}, which indicates the complexity of the AES task. These results show the critical role of prompt engineering in enhancing or diminishing the performance of LLMs on the AES task. Moreover, choosing the right prompt is highly dependent on \emph{both} the LLM and the nature of the task.

This also highlights that, despite the astonishing ability of LLMs to generate coherent and good-quality text, they struggle to distinguish between good and bad essays. This issue persists even when providing more context about the scores and a sample of a scored essay.

\begin{table*}[]
\centering
\begin{tabular}{ccccccccc}
\hline
\multirow{2}{*}{Task} & \multirow{2}{*}{LLM} & \multicolumn{6}{c}{Prompt Pairs} & \multirow{2}{*}{Average} \\
  \cmidrule{3-8} &  & A - B & A - C & A - D & B - C & B - D & C - D &  \\ \hline
1 & ChatGPT & 0.772 & 0.526 & 0.449 & 0.672 & 0.585 & 0.741 & 0.62 \\
 & Llama & 0.286 & 0.595 & 0.615 & 0.235 & 0.293 & 0.537 & 0.43 \\ \hline
5 & ChatGPT & 0.836 & 0.709 & 0.458 & 0.734 & 0.405 & 0.312 & 0.58 \\
 & Llama & 0.059 & 0.046 & 0.222 & 0.524 & 0.181 & 0.162 & 0.20 \\ \hline
7 & ChatGPT & 0.804 & 0.748 & 0.477 & 0.756 & 0.475 & 0.607 & 0.64 \\
 & Llama & 0.457 & 0.148 & 0.152 & 0.092 & 0.125 & 0.021 & 0.17 \\ \hline
\end{tabular}%
\caption{QWK performance of each prompt pair for ChatGPT and Llama}
\label{tab:consistency_evaluation}
\end{table*}

\subsection{LLMs vs. SOTA (RQ3)} \label{sec:llmVsSOTA}
\label{rq2}

\begin{table*}[h]
\centering
\begin{tabular}{ccc|cc}
\hline
Task & \citet{xie2022automated}  & \citet{jiang-etal-2023-improving}  & ChatGPT & Llama \\ \hline
1 & \textbf{0.856} & 0.762 & 0.120 & 0.331 \\
2 & \textbf{0.750} & 0.686 & 0.193 & 0.562 \\
3 & \textbf{0.756} & 0.637 & 0.198 & -0.059 \\
4 & \textbf{0.851} & 0.673 & 0.416 & 0.201 \\
5 & \textbf{0.847} & 0.778 & 0.576 & 0.418 \\
6 & \textbf{0.858} & 0.664 & 0.606 & 0.363 \\
7 & \textbf{0.838} & 0.742 & 0.123 & 0.359 \\
8 & \textbf{0.779} & 0.677 & 0.276 & 0.201 \\ \hline
Average & \textbf{0.817} & 0.702 & 0.313 & 0.297 \\ \hline
\end{tabular}%
\caption{QWK performance of ChatGPT and Llama vs. SOTA holistic models. Best per row is boldfaced.}
\label{tab:holistic_evaluation}
\end{table*}

We next turn to measure the performance of the two LLMs in predicting the holistic score and trait scores of the essays over the entire set of ASAP tasks. With that, we aim to (1) comprehensively contrast the performance of the two LLMs against each other, and (2) position their performance within the best AES models in the literature.

Based on the first stage, for each LLM, we selected the prompt that achieved the highest QWK for each task category. Specifically, for ChatGPT, we selected prompts C, D, and B, for the persuasive, narrative, and source-dependent categories, respectively. Likewise, for Llama, we opted for prompts A, B, and C, for the persuasive, narrative, and source-dependent categories, respectively.

Table~\ref{tab:holistic_evaluation} compares the performance of ChatGPT and Llama per ASAP task with SOTA models (as reported in their respective studies) on the holistic level. 
First, we notice that the average performance of the two LLMs across tasks is quite comparable, with a slight advantage for ChatGPT.
The results also show that ChatGPT does a reasonably good job in source-dependent tasks 4-6, where the performance is significantly better than in the other (source-independent) tasks, while Llama showed a better performance only in Task 2 (which is persuasive).
Second, it is quite apparent that the exhibited performance of both LLMs is nowhere near the SOTA models, with average QWK of 0.817, 0.702, 0.313, and 0.297 for \citet{xie2022automated}, \citet{jiang-etal-2023-improving}, ChatGPT, and Llama, respectively. 
While those SOTA models are task-specific and trained on the ASAP dataset directly,  higher performance was expected from the LLMs, given the vast amount of data they were trained on and their ability to generate good-quality text. It is clear that although LLMs were able to understand the task and give a score, it was not able to capture the input essay as a whole and score it accordingly.

\begin{table*}[h]
\begin{tabular}{ccccccccc}
\hline
Task & Model & Content & Organization & Voice & Word   Choice & SF & Conventions & Style \\ \hline
7 & BEA 2020 & 0.771 & 0.676 & - & - & - & 0.621 & 0.659 \\
 & ChatGPT & 0.045 & 0.068 & - & - & - & 0.097 & 0.079 \\
 & Llama & 0.091 & 0.023 & - & - & - & 0.327 & 0.154 \\ \hline
8 & BEA 2020 & 0.586 & 0.632 & 0.544 & 0.559 & 0.586 & 0.558 & - \\
 & ChatGPT & 0.185 & 0.256 & 0.155 & 0.15 & 0.203 & 0.311 & - \\
 & Llama & 0.276 & 0.271 & 0.278 & 0.268 & 0.121 & 0.089 & - \\ \hline
\end{tabular}%
\caption{Comparison of ChatGPT's and Llama's performance against \citet{mathias_bhattacharyya_2020_neural} (BEA 2020) per trait scores across tasks 7 \& 8. 
SF refers to Sentence Fluency.}
\label{tab:trait_evaluation}
\end{table*}

On the traits level, Table~\ref{tab:trait_evaluation} draws a comparison with SOTA models per trait for tasks 7 and 8, the only tasks that have traits in ASAP. The results deliver the same message of the holistic level; ChatGPT and Llama are very far behind.

\subsection{Analysing the Provided Feedback}
Despite the underwhelming performance of both LLMs in scoring essays, that does not mean that LLMs cannot help in sharpening the essay quality by providing feedback for the writer. Here, we show a sample of such feedback and discuss the difference between ChatGPT and Llama. Table~\ref{tab:feedback} shows the feedback of both LLMs on two sample essays from Task 3 (source-dependent), one is of low quality and the other is of high quality.

Interestingly, for ChatGPT, the feedback provided for the low-score essay was short, and there was no elaboration or meaningful comments on how to improve the essay. On the contrary, Llama elaborated better on why that essay had a low score. For the high-score essay, both models were able to explain where the essay could be improved; however, again, Llama was able to better explain where the essay could be improved. 

While this is a very small sample (only two essays), it shows promising potential for LLMs to provide meaningful feedback on the writing quality of the given essays, which indeed opens a potential research direction for future work on using LLMs in the AES task. 
We plan to evaluate a larger set of feedback responses in future experiments to further validate and extend these findings.

\begin{table*}[t]
    \centering
    \footnotesize
    \begin{tabular}{|p{1.8cm}|p{5.cm}|p{8.0cm}|}
        \hline
        Essay & ChatGPT & Llama \\
        \hline
        5985 \newline (Task 3, \newline \textbf{Low-Score}) & 
        The response is completely irrelevant and incorrect. It does not address the prompt and does not demonstrate any understanding of the text. &
        The response shows evidence of a minimal understanding of the text. The essay does not address the prompt or the features of the setting that affect the cyclist. Instead, the essay focuses on the writer's personal feelings and experiences, which are not related to the prompt. The essay also contains some inaccuracies, such as the statement that the cyclist was "cut hurt" and that the features of the setting did not affect the cyclist. Overall, the response lacks a clear understanding of the text and the prompt. \\
        \hline
        7000 \newline (Task 3, \newline \textbf{High-Score}) & 
        The response demonstrates a partial understanding of the text. The essay discusses how the features of the setting affect the cyclist, specifically how the old directions and abandoned towns with no good water could have killed him. However, the response could have provided more examples from the essay to support the conclusion. Additionally, the response could have extended the understanding beyond the literal by discussing how the cyclist's experience reflects the fluidity of the world and the importance of being prepared for nature's challenges. &
        The response demonstrates a partial understanding of the text, although may not develop all parts equally. The essay provides some examples from the text to support the conclusion, but the analysis is not comprehensive and does not fully connect the support to a conclusion or assertion made about the text. The response correctly identifies the features of the setting that affect the cyclist, such as the desert environment, the lack of water, and the abandoned towns. However, the analysis could be more detailed and nuanced. For example, the response could discuss how the setting affects the cyclist's mood and confidence, or how the abandoned towns and lack of water create a sense of desolation and hopelessness. Additionally, the response could provide more specific examples from the text to support the analysis. Overall, the response demonstrates some understanding of the text, but could be more comprehensive and nuanced in its analysis. \\
        \hline
    \end{tabular}
    \caption{Feedback generated by ChatGPT and Llama for a sample essay of actual low and high scores.}
    \label{tab:feedback}
\end{table*}

\section{Limitations}

Our study has indeed several limitations. 
\begin{itemize}
    \item First, in our experiments, we were keen on making the length of the prompt as short as possible to fit within the maximum-allowed limit of tokens (which is 4,096 tokens) and reduce the corresponding cost. 
    \item Second, we \emph{randomly} chose one example for the one-shot prompts and did not experiment with different examples. It is also possible that adding few-shot examples might achieve better results. 
    \item A further limitation is that OpenAI does not keep track of the historical context of previous requests. So, we had to send the same evaluation instructions in each API request, which incurred additional costs. 
    \item Finally, we experimented with one closed- and another open-source LLMs. Experimenting with more LLMs might remove some of the above limitations and bring additional insights.
\end{itemize}

\section{Conclusion} \label{conclusion}

In this study, we examined the ability of two LLMs, namely ChatGPT and Llama, to assess written essays holistically and per trait, analyzed their performance, and studied their consistency. 
We designed 4 prompts that showed different performance over different task categories. Our analysis shows that the performance of the LLMs is highly dependent on the prompt and the task type. Moreover, Llama, in particular, is very sensitive to small changes in the prompt, while ChatGPT is more robust and consistent. 
Although both LLMs fell short behind other SOTA models, it is inevitable to say that LLMs are high-quality text-generating machines. However, it is important to highlight their limitations when it comes to~\emph{evaluating} text. Our study suggests a possibility of utilizing the vast linguistic knowledge those LLMs have to provide feedback for improving essays. However, according to our study, LLMs are not yet reliable for predicting the score a student would get on an essay.

In future work, we plan to analyze the generated feedback and its usefulness, besides examining the performance of other LLMs, such as Bard, on the task. Expanding our study to other datasets is another future direction. Fine-tuning the LLMs for the AES task might also improve the performance; therefore, that direction is also worth exploring. 

\section{Acknowledgements}

The work of Salam Albatarni was supported by GSRA grant\# GSRA10-L-2-0521-23037 from the Qatar National Research Fund (a member of Qatar Foundation). The work by the rest of the authors was made possible by NPRP grant\# NPRP14S-0402-210127 from the Qatar National Research Fund. The statements made herein are solely
the responsibility of the authors.

\nocite{*}
\section{Bibliographical References}\label{sec:reference}

\bibliographystyle{lrec-coling2024-natbib}
\bibliography{lrec-coling2024-example}

\clearpage

\end{document}